%% file: main.tex
\pdfoutput=1 
\documentclass[10pt,twocolumn,a4paper]{article}

\usepackage{times}
\usepackage{epsfig}
\usepackage{graphicx}
\usepackage{subfig}
\usepackage{amsmath}
\usepackage{amssymb}
\usepackage{stfloats}
\usepackage{svg}
\usepackage{import}
\usepackage[nolist]{acronym}
\usepackage{enumitem}
\usepackage{hhline}
\usepackage[a4paper, lmargin=1.8cm, rmargin=1.8cm, top=20mm, bottom=25mm]{geometry}
\usepackage[misc]{ifsym} %
\PassOptionsToPackage{%
	backend=bibtex,%
	bibstyle=ieee, %
	giveninits=false, %
	url=false,%
	isbn=false,%
	doi=false,%
	hyperref=auto,%
	natbib=true, %
}{biblatex}
\usepackage{biblatex}
\bibliography{main}

\usepackage{hyperref}

\newcommand{\ie}{i.e.\ }

\newcommand{\eg}{e.g.\ }

\newcommand{\wrt}{w.r.t.\ }

\begin{document}

\title{Exploiting the Full Capacity of Deep Neural Networks while Avoiding Overfitting by Targeted Sparsity Regularization}

\author{Karim Huesmann\footnote{both authors contributed equally}\ , S\"oren Klemm\footnotemark[1]\ , Lars Linsen, Benjamin Risse\footnote{\Letter\ b.risse@uni-muenster.de} \\
Westf\"alische Wilhelms-Universit\"at M\"unster\\
M\"unster, Germany\\
}

\date{}
\maketitle

\input{section/abstract.tex}

\input{section/introduction.tex}
\input{section/related.tex}
\input{section/methodology.tex}
\input{section/experiments.tex}

\input{section/conclusion.tex}
{
\printbibliography
}
\appendix
\input{section/appendix}
\end{document}

%% file: section/abstract.tex
\begin{abstract}
\noindent Overfitting is one of the most common problems when training deep neural networks on comparatively small datasets. 
Here, we demonstrate that neural network activation sparsity is a reliable indicator for overfitting which we utilize to propose novel targeted sparsity visualization and regularization strategies.
Based on these strategies we are able to understand and counteract overfitting caused by activation sparsity and filter correlation in a targeted layer-by-layer manner. 
Our results demonstrate that targeted sparsity regularization can efficiently be used to regularize well-known datasets and architectures with a significant increase in image classification performance while outperforming both dropout and batch normalization.
Ultimately, our study reveals novel insights into the contradicting concepts of activation sparsity and network capacity by demonstrating that targeted sparsity regularization enables salient and discriminative feature learning while exploiting the full capacity of deep models without suffering from overfitting, even when trained excessively. 
\end{abstract}

%% file: section/introduction.tex
\section{Introduction}

\noindent Deep NNs have achieved remarkable success in a variety of different application domains such as Computer Vision~\cite{he2017mask,ronneberger2015unet,teichmann2018multinet}, Pattern Recognition~\cite{saufi2019challanges,shone2018deep}, or Natural Language Processing~\cite{anderson2018bottom,devlin2018bert}.
The training success of NN models mainly relies on three fundamental requirements: (i) the amount of available training data; (ii) the theoretical capacity of the trainable model; and (iii) the learning strategy including the hyperparameters.
All requirements are interdependent and have to be considered as a whole.

The capacity of the model is mainly defined by its architecture, which has to be big enough to learn potentially complex data relations without causing an overadaption to the training data resulting in poor generalization capabilities.
Given inappropriate combinations of network size and training data special measures have to be considered to prevent overfitting~\cite{zhang2017understanding}.
Popular countermeasures include (i) artificially increasing the amount of training data~\cite{cubuk2019autoaugment,guo2019mixup}; (ii) reducing the network's capacity; and (iii) changing the learning strategy~\cite{yaguchi2018adam}. 
The reduction of the capacity can either be done explicitly by reducing the amount of learnable parameters (\ie pruning)~\cite{goodfellow2016deep,liu2017learning,klemm2019deploying} or implicitly by regularizing the trainable parameters of a model~\cite{nowlan1992simplifying,Ioffe2015}.
While a reduction of the NN's capacity can lead to less powerful networks, synthetically generated training data can induce a variety of side effects~\cite{geirhos2018imagenet}.
Different training strategies do not change the network's capacity and training data.
However, the effect of these strategies is yet not completely understood so that they still require a trial-and-error usage~\cite{reddi2018convergence}.

In this work, we propose a different approach to tackle overfitting, namely targeted sparsity regularization.
In contrast to existing work in which sparsity is often considered a desirable network property~\cite{ide2017improvement,papyan2018theoretical}, we demonstrate that sparse activations in intermediate layers are in fact a reliable indicator for overfitting.
By exploiting this concept, we derive a novel visualization strategy to diagnose overfitting in individual layers of convolutional neural networks during training.
Furthermore, these insights are used to introduce a targeted per-layer regularization strategy which avoids the under-utilization of the network's capacity while increasing the overall performance.
Moreover, we demonstrate how targeted sparsity regularization enables to train larger NN architectures on comparatively small datasets while preventing overfitting entirely even when trained excessively.
As sparsity also limits the amount of information stored in a given number of activations~\cite{pezzotti2018deepeyes}, we furthermore demonstrate that the predictive power of a NN with a fixed size increases when information from multiple extracted features are combined.

In summary we present four central contributions:
\begin{enumerate}[label=C\arabic*:]
    \item We provide an interactive visualization method based on the sparsity of activations that enables the identification of overfitting on a per-layer basis which can directly be embedded into TensorBoard.
    \item We introduce a novel targeted regularization strategy exploiting the sparsity of the activations in combination with decorrelated convolutional filters, which can prevent overfitting even for very long trainings.
    \item We demonstrate that this regularization strategy \textbf{}significantly increases image classification performance on well-known datasets using different NN architectures while outperforming both dropout and batch normalization.
    \item We provide novel insights into the seemingly contradicting concepts of activation sparsity versus network capacity by demonstrating that deep NNs can be regularized in order to learn distinctive and salient features without inducing low or redundant activations.
\end{enumerate}

%% file: section/related.tex
\section{Related Work}\label{ch:related}
\noindent
Work on regularizing overfitting can be separated into two different categories, namely topology-based and loss-based regularization.
As the name suggests, topology-based regularization changes the neural connections or incorporates additional layers into the network's architecture. 
The two most common examples are dropout~\cite{Srivastava2014} and batch normalization~\cite{Ioffe2015}.
Dropout layers temporarily switch off random neurons during the training process, inducing a noisy input to the subsequent (hidden) layers~\cite{Srivastava2014}. 
In contrast, batch normalization layers were initially designed to reduce the internal covariate shift to accelerate the training process~\cite{Ioffe2015}. 
Since the normalization is initialized by subtracting the mean and dividing by standard deviation of the current batch activations, this also induces noise to subsequent layers which results in regularization side-effects.
The common principle of those methods is to artificially modify the training input and hence reduce the chance of overadaption.
While this approach has proven to be effective in many applications, the introduced noise can also prevent optimal performance, as can be seen when combining dropout and batch normalization, which in general leads to degraded performances~\cite{Li2019,Klambauer2017}.
 
In loss-based regularization strategies neither the training input nor the network's architecture is changed.
Instead, the loss is altered by an explicit regularization term to cope with overfitting. 
Loss-based regularization can again be separated into two common paradigms, differing in the domain of the regularizer, which can either be the hidden activations / weights or the output distribution. 
Based on the observation that large deep neural networks often lead to redundancies~\cite{ayinde2019correlation}, many of the activation- / weight-based strategies aim to extenuate the model complexity by using weight decay~\cite{nowlan1992simplifying,ayinde2017deep} or pruning the network directly~\cite{molchanov2016pruning,liu2017learning}.
Similarly, it has been shown that classical L1 regularization induces sparsity while L2 regularization causes the decay of weights over time~\cite{yaguchi2018adam,mehta2019implicit}.
Even though these techniques can improve the generalizability and performance of the network, they underutilize the potential capacity of the model~\cite{ayinde2019regularizing}.

In contrast, decorrelation-based regularizers aim to improve the networks while attempting to employ the given capacity.
Both hidden features~\cite{bengio2009slow} and activations~\cite{bao2013incoherent,cogswell2015reducing} are utilized to reduce the redundancy within the model.
Others try to avoid correlations by reducing the cosine similarity among feature vectors to avoid overfitting while improving the overall performance~\cite{ayinde2019regularizing}.

In the second category of loss-based regularization, the output distribution of the network is used to achieve less overfitting. 
In particular maximum entropy-based regularization has long been studied to regulate the models behavior~\cite{miller1996global}. 
In a more recent approach Pereyra et al. propose to penalize the entropy of highly confident Softmax outputs~\cite{pereyra2017regularizing}. 
However, since the entropy is derived from the predictions of the last layer, this regularization does not directly target the layers, which cause the loss of information.

In fact, all the optimization strategies mentioned above have in common that they regularize the network in a non-targeted manner:
Instead of identifying the layers which are responsible for overfitting, the regularization is applied to either all layers (\eg ecorrelation-based regularization) or to randomly selected entities within the network (\eg dropout).
Furthermore, none of the above-mentioned techniques explicitly address the contradicting trade-off between sparse but salient activations (\ie sparsity) and low or redundant feature responses (\ie network capacity).
In contrast our proposed method allows a targeted regularization and decorrelation of layers thus avoiding overfitting of the NN.

%% file: section/methodology.tex
\section{Methods}
\subsection{Measuring Sparsity}
\noindent Sparse activations in CNNs are inspired by mammalian visual cortex cells and have been studied extensively for feature extraction purposes~\cite{Olshausen1996}.
Based on the idea that each filter of a convolutional layer is trained to identify a particular feature of a given input~\cite{Zeiler2014}, this work analyzes the perplexity~\cite{pezzotti2018deepeyes} of receptive field activations to quantify sparsity (\textbf{C1}). %
The main reason for analyzing perplexity is to determine whether a filter is trained in such a way that it returns an overconfident output distribution for a given receptive field. 

\begin{figure*}
	\centering 
	\includegraphics[width=\linewidth]{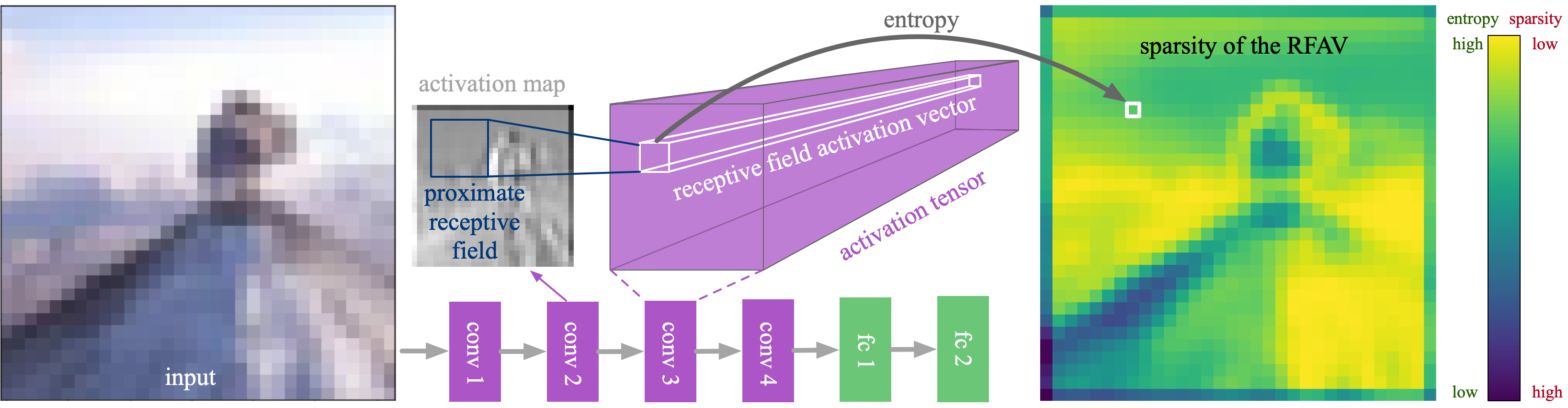}
	\centering
	\caption{Heatmap visualization of sparsity:
	Left: sample input image.
	Middle: used architecture and sketch of an exemplary proximate receptive field from the conv 2 layer and the corresponding receptive field activation vector (RFAV) in conv 3 layer. 
	Right: sparsity heatmap visualizations are extracted by calculating the entropy on the RFAVs. Note that a high entropy corresponds to a low sparsity and vice versa.}
	\label{fig:sparsity_visualization}
\end{figure*}

In this work, the \textit{proximate receptive field} is defined as the region of a layer's direct input that a filter is being affected by (see \autoref{fig:sparsity_visualization}). 
This definition differs from the conventional definition of receptive fields which usually define the area in the network's input space. We will further focus on 2D convolutions, common for image data. 
All reasoning, however, also applies to other dimensionalities and dense layers.

A layer with $D$ filters may create a $D \times A \times B$ shaped feature tensor. 
Thus, the input of this layer consists of $A \cdot B = R$ receptive fields. 
Let $x_{d',i,j,a,b}$ be a pixel at position $(i, j)$ and channel $d'$ of receptive field $r_{a,b}$ .
The corresponding weight of filter $f_d$ which affects this pixel is given as $w_{d',i,j,d}$.
The number of pixels in $r_{a,b}$ is equal to the number of weights in $f_d$.
Hence, the linear activation $a_{a,b,d}$ created by $f_d$ applied to $r_{a,b}$ is defined as
\begin{equation}
a_{a,b,d} = \sum_{d',i,j} w_{d',i,j,d} \cdot x_{d',i,j,a,b} \ .
\end{equation}

To improve readablity, we denote the results of this linear filtering as receptive field activation vectors (RFAV) $\textbf{a}_k \in \mathbb{R}^D, k \in [0,R-1]$, where $k$ is a linear index over all receptive fields computed by $k = a \cdot B + b$ (see \autoref{fig:sparsity_visualization}).

We encode the sparsity of receptive field activations with the help of perplexity. Perplexity $\rho$ is defined as 
\begin{equation}
    \rho(H(\textbf{p})) = e^{H(\textbf{p})}
\end{equation}
where $\textbf{p}$ is a discrete probability distribution and $H(\textbf{p})$ the corresponding entropy of $\textbf{p}$. 
Since perplexity is strictly monotonically increasing \wrt entropy, our approach focuses on entropy only.  
In order to calculate the entropy of a RFAV $\textbf{a}_k$, its components have to be transformed into a probability-like distribution. 
This can be done by applying the Softmax function:
\begin{equation}
p_l(\textbf{a}_k) = \frac{e^{a_k^l}}{\sum_m e^{a_k^m}} \ , \ l \in [1, ..., D] 
\end{equation} 
where $a_k^l$ is the $l$-th component of $\textbf{a}_k$. 
For every receptive field we obtain the corresponding RFAV entropy by 
\begin{equation}
\label{eq:entropy}
H_k = H(\textbf{a}_k) = - \sum_{l=1}^D p_l(\textbf{a}_k) \ ln(p_l(\textbf{a}_k)) \ .
\end{equation}
The values of $H_k$ correlate with the respective sparsities, such that sparse activations result in RFAV entropy values close to zero and dense activations in large values. %
Encoding and re-ordering all $H_k$ of a layer in a 2D heat map allows to visualize the localized sparsity of activations for a given input (see \autoref{fig:sparsity_visualization} and contribution \textbf{C1}).

\subsection{Regularizing Sparsity} \label{sec:methods:regularizing-sparsity}
\noindent As we will show in  \autoref{sec:monitoring_sparsity}, when overfitted, neurons (\ie filters) in a NN have been adapted to particular features and only cover individual observations including noise and fluctuations. 
Thus, sparse activations become much more common when a neural network overfits~(see \autoref{fig:mean-entropy-xxl-unreg-drpt-bn}).
Instead of preventing overfitting by artificially modifying training data, we propose a penalty term that prevents the generation of sparse activations directly (\textbf{C2}).

As entropy is differentiable, it can be used as an activity regularizer to control RFAV sparsity. 
In order to regularize sparsity, the loss function is extended by a penalty term
\begin{equation}
\label{eq:sparsity-reg}
\mathcal{L}_{s} = - \sum_{i} \; \lambda_i \sum_{k=0}^{r_i}  H_{k}^{i} ,
\end{equation}
where $i$ loops through all layers of the NN, $r_i$ is the number of receptive fields in the respective layer and $H_{k}^{i}$ refers to $H_k$ as defined in \autoref{eq:entropy} \wrt layer $i$. In the following this regularization is referred to as \textit{SparsityReg}.
Furthermore, $\lambda_i \geq 0$  can be used to toggle the sparsity regularizer of layer $i$.

As entropy reaches its maximum when all filters are activated in the same way, this regularizer can have the tendency to produce highly correlated filters. 
The trivial solution for $\mathcal{L}_{s}$ to be minimized is therefore to generate identical filter responses.
As identical filters reduce the predictive power of NNs, this effect has to be counterbalanced by preventing high filter correlations.

\subsection{Regularizing Filter Correlations}
\noindent Considering a layer with $D$ filters (or neurons), each filter consists of weights $\textbf{w}_d=(w_{d',i,j})$.
If the weights of a neuron $d$ strongly correlate (or anti-correlate) with the weights of another neuron $e$, they create redundant feature maps. 
The correlation coefficient $c_{d,e}$ of these neurons is calculated as follows (Pearson correlation):

\begin{equation} \label{eq:weight-corr}
	c_{d,e} = \frac{(\textbf{w}_d - \bar{\textbf{w}}) \cdot ( \textbf{w}_e - \bar{\textbf{w}})}{\sqrt{(\textbf{w}_d - \bar{\textbf{w}})^2 \cdot (\textbf{w}_e - \bar{\textbf{w}})^2}} 
\end{equation} 
with $\bar{\textbf{w}}$ being a vector holding the mean of all $\textbf{w}_{d}$'s in each component.
Calculating pairwise correlation coefficients of all filters in a layer results in a correlation matrix. 
Correlations can be visualized using a 2D histogram (see \autoref{fig:correlation-histogram}). 
Here, the $x$-axis corresponds to the epoch, $y$-axis to the correlation coefficient. Due to symmetry it suffices to analyze the lower triangular matrix of the correlation matrix.
Color encodes the frequency how often a correlation coefficient appears in the lower triangle of the correlation matrix for the corresponding epoch. 
In this visualization one can observe  how the correlation coefficients are distributed and how this distribution changes in the course of training.

Since Pearson correlation coefficient is a continuous differentiable function, a correlation regularizer can be added to the training loss in order to prevent high correlations (\textbf{C2}). 
The correlation regularizer $\mathcal{L}_{c}$ is given as
\begin{equation}
	\mathcal{L}_{c} = \sum_i \kappa_i \sum_d \sum_{e>d} c_{d,e}
\end{equation}
where $\kappa_i \geq 0$ is used to control the strength of correlation regularization of layer $i$. In the following this regularization is referred to as \textit{DecorrReg}.
The overall loss used for training can now be written as:
\begin{equation}
    \mathcal{L} = \mathcal{L}^\ast + \mathcal{L}_s + \mathcal{L}_c ,
\end{equation}
where $\mathcal{L}^\ast$ denotes the loss used to measure inference quality (\eg categorical cross-entropy loss).

%% file: section/experiments.tex
\begin{figure}[htb]
	\centering 
    \includegraphics{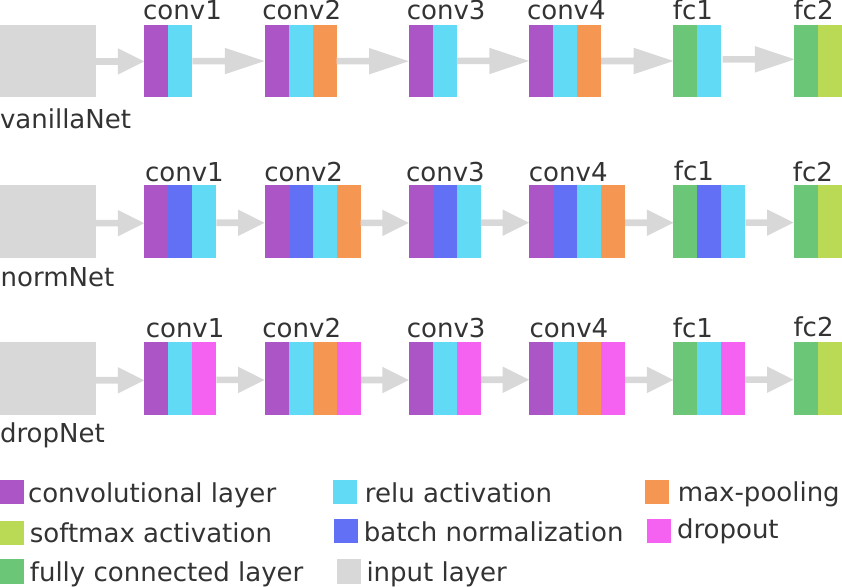}
	\caption{Schematic representation of our baseline architecture. Colors encode the different layer types. The convolutional and max-pooling layer refer to 2D convolution and 2D max-pooling.}
	\label{fig:network}
\end{figure}
\section{Experiments}
\noindent First, we evaluate our method using a common baseline architecture depicted in \autoref{fig:network}. 
The basic network comprises two conv-conv-pool blocks followed by two dense (or fully connected) layers. 
ReLU is used as activation across all layers.
In the following, this network is referred to as \textit{vanillaNet}. 
A second architecture, where every conv block is extended by a batch normalization layer, we refer to as \textit{normNet}.
A vanillaNet with added dropout ($p=0.25$) in every conv block is called \textit{dropNet}.
Information on different architecture sizes can be found in \autoref{table:network}. 
If not stated otherwise, categorical cross-entropy loss and SGD optimizer with learning rate of 0.01 and no momentum are used for evaluation. 
The networks are trained for 100 epochs with a batch size of 64. 
During training, a total of 1,000 samples, which are neither part of the training nor of the validation dataset, are used to calculate the layer's RFAV sparsity. 
In total, three datasets (cifar-10, cifar-100, tiny-imagenet) are considered for training.
Throughout all training runs' $\kappa_i$ were set to either $0$ or $1$ to switch correlation regularization of the respective layer off or on.
Sparsity regularization was controlled by setting $\lambda_i$ to either $0$ or $0.001$. This value was chosen manually and offered favorable results throughout our experiments. 
\begin{table}
    \centering
	\begin{tabular}{lccccc}
		\textbf{Size} & \textbf{conv1} & \textbf{conv2 } &\textbf{conv3 } & \textbf{conv4 } & \textbf{fc1} \\ \hline
		s   &  16 &  16 &  32 &  32 &   64 \\
		m   &  64 &  64 & 128 & 128 &  256 \\
		xxl & 256 & 256 & 512 & 512 & 1024                                 
	\end{tabular}
	\caption{Overview of number of filter channels / neurons for different sizes of evaluated networks.}
	\label{table:network}
\end{table}

\subsection{Monitoring Layer Sparsity}
\label{sec:monitoring_sparsity}
\begin{figure*}
    \subfloat[\label{fig:loss-small}]{
        \begin{minipage}[b]{0.33\textwidth}
           \includegraphics[width=\linewidth]{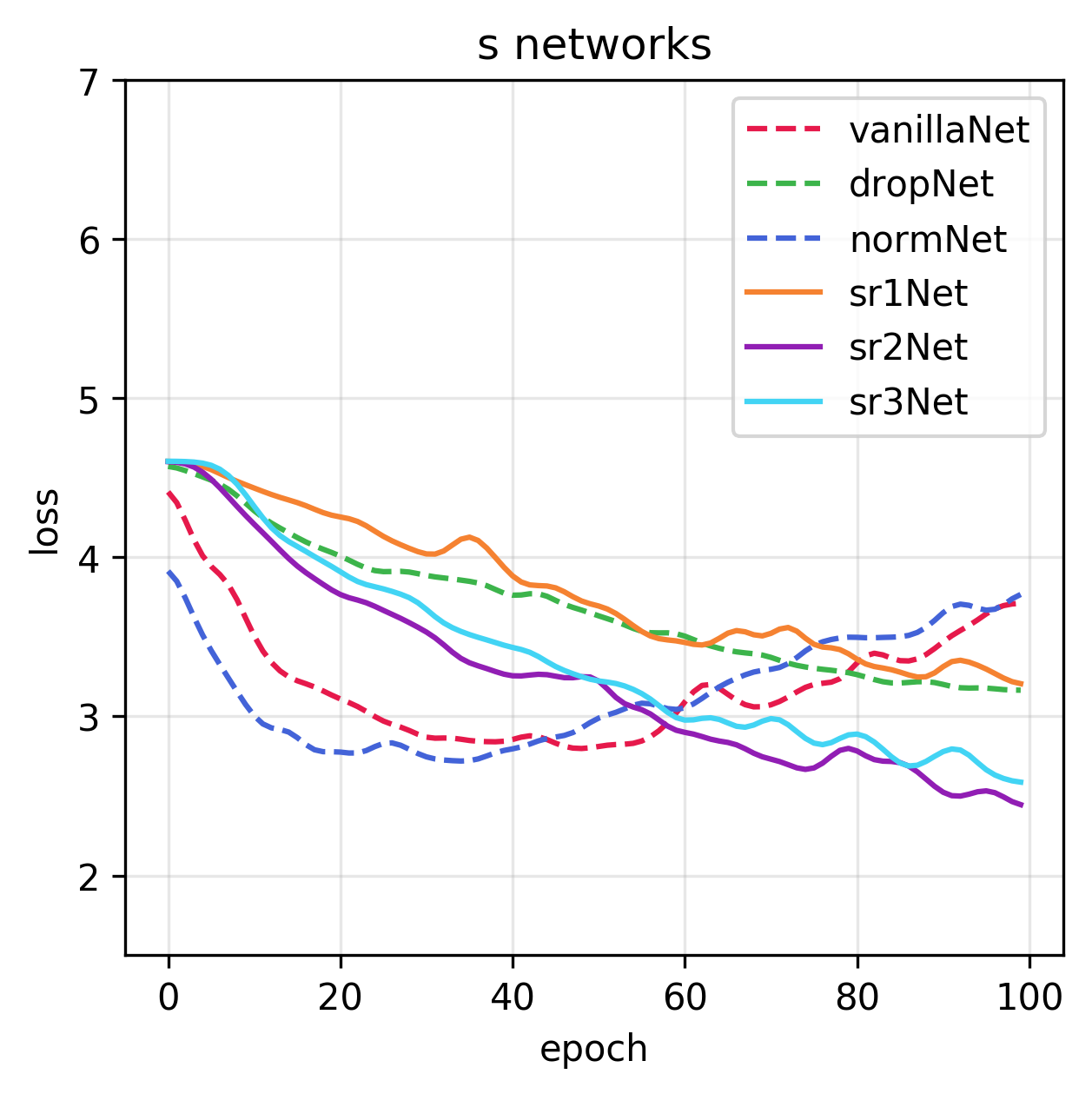}
        \end{minipage}
    }
    \begin{minipage}[b]{0.33\textwidth}
        \subfloat[\label{fig:loss-medium}]{
            \begin{minipage}[b]{\textwidth}
                \includegraphics[width=\textwidth]{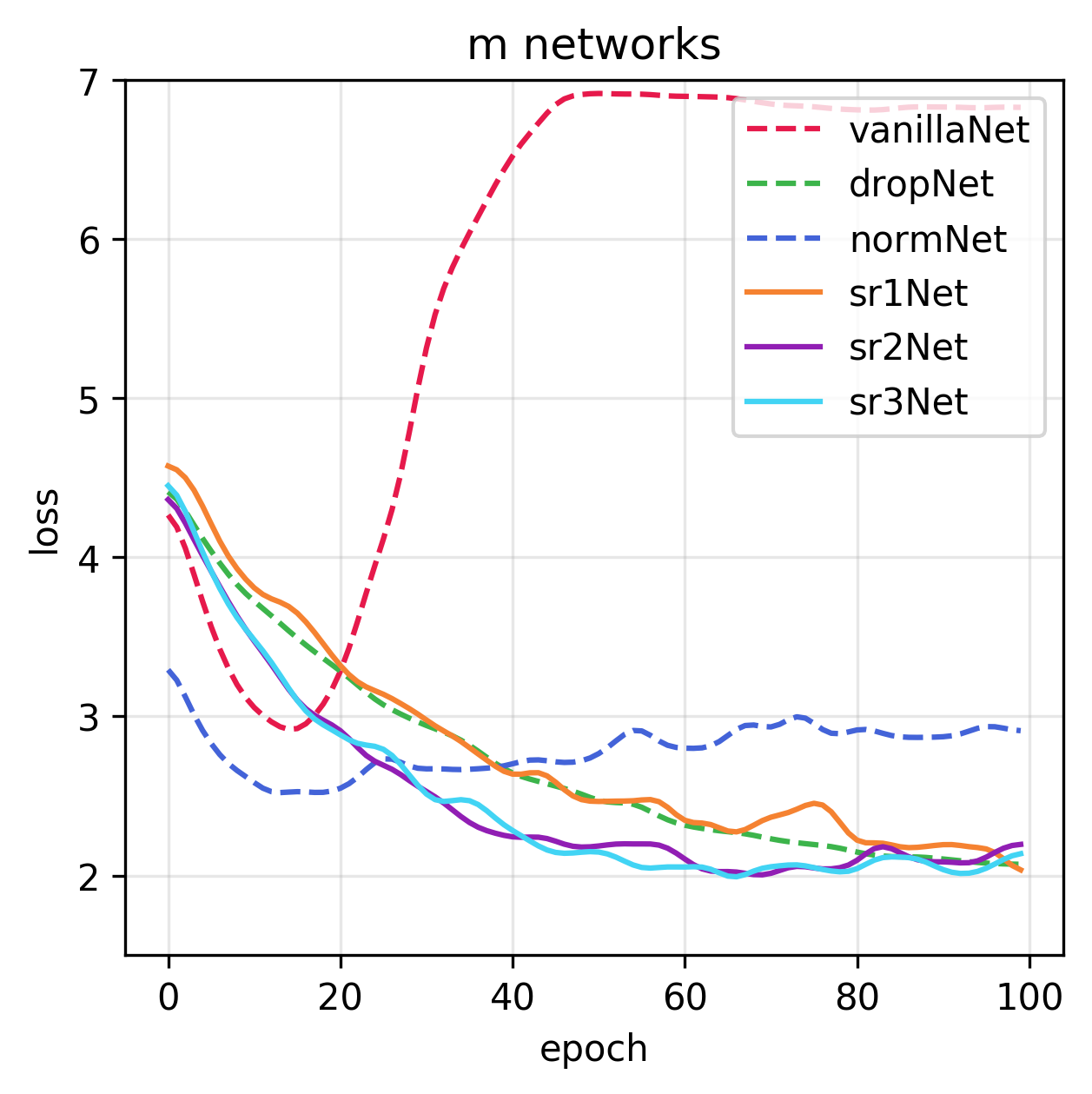}
            \end{minipage}
        }
    \end{minipage}
    \begin{minipage}[b]{0.33\textwidth}
        \subfloat[\label{fig:loss-large}]{
            \begin{minipage}[b]{\textwidth}
                \includegraphics[width=\textwidth]{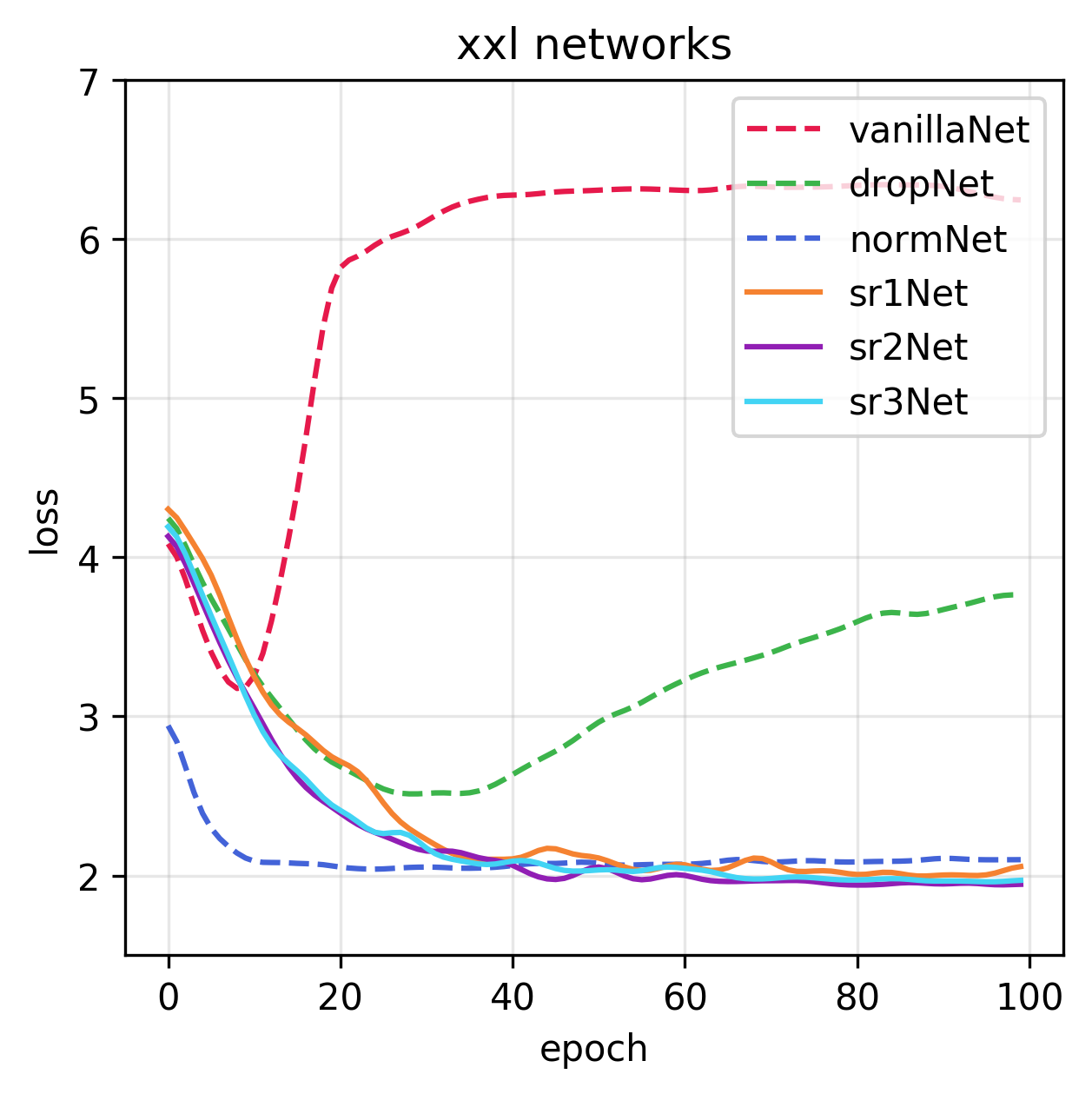}
            \end{minipage}
        }
    \end{minipage}
\caption{Evolution of categorical cross-entropy loss of small (a), medium (b), and xxl-sized (c) networks when trained on cifar-100.
A gaussian kernel ($\sigma$ = 2 epochs) was used for smoothing. For a better comparability, the regularized losses were adjusted by their penalty terms.
} 
\label{fig:loss-all}
\end{figure*}
\begin{figure*}

    \subfloat[\label{fig:entropy-heatmap:unreg}]{
    \begin{minipage}[b]{0.49\textwidth}
       \includegraphics[width=\linewidth]{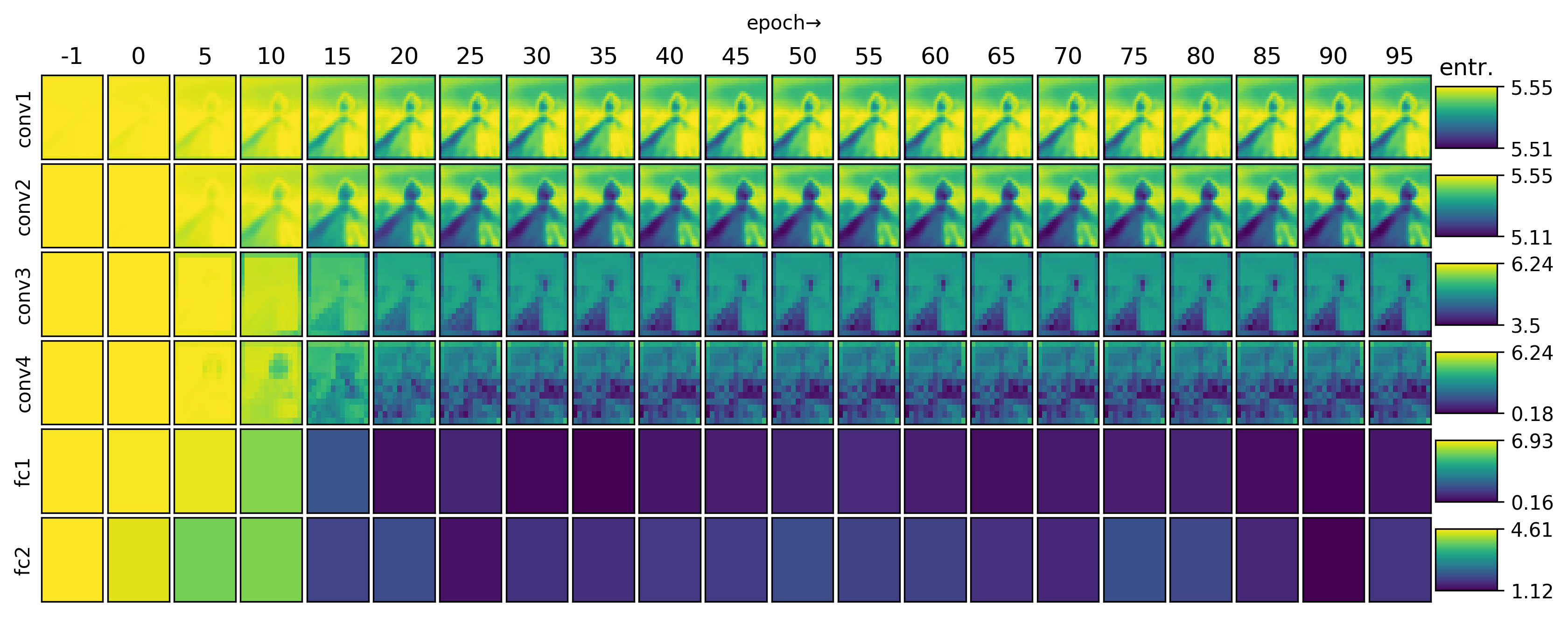}
       \end{minipage}
    }
    \hfill
    \hfill
    \begin{minipage}[b]{0.49\textwidth}
        \subfloat[\label{fig:entropy-heatmap:drpt}]{
            \begin{minipage}[b]{\textwidth}
                \includegraphics[width=\linewidth]{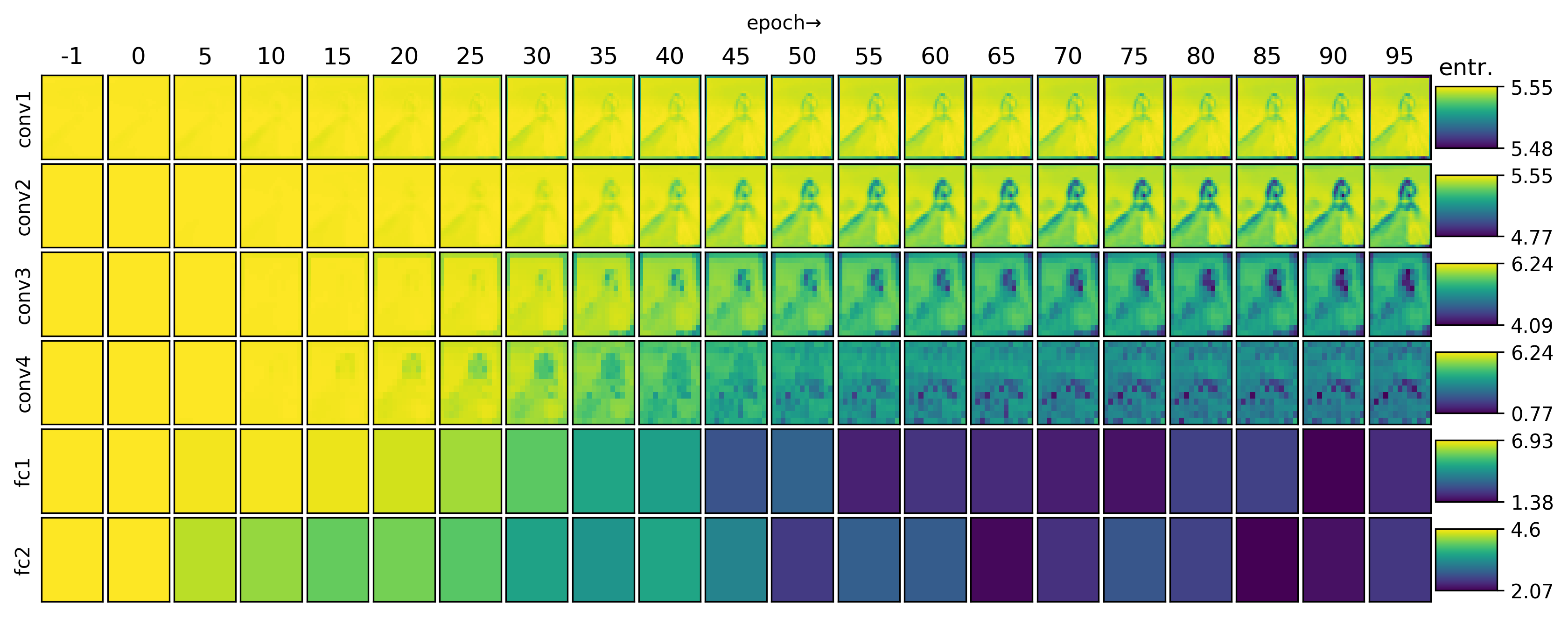}
            \end{minipage}
        }
    \end{minipage}
    \subfloat[\label{fig:entropy-heatmap:bn}]{
    \begin{minipage}[b]{0.49\textwidth}
        \includegraphics[width=\linewidth]{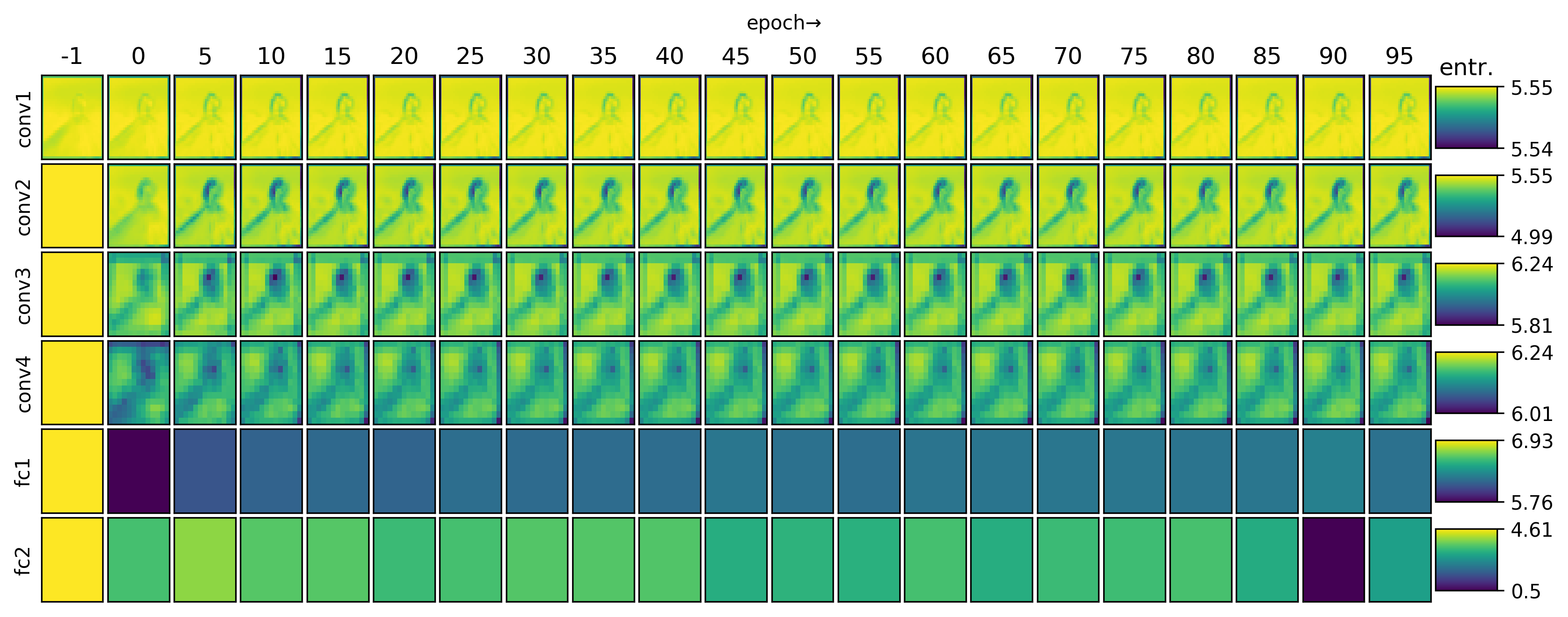}
        \end{minipage}
    }
    \hfill
    \hfill
    \begin{minipage}[b]{0.49\textwidth}
        \subfloat[\label{fig:entropy-heatmap:sparsereg}]{
            \begin{minipage}[b]{\textwidth}
                \includegraphics[width=\linewidth]{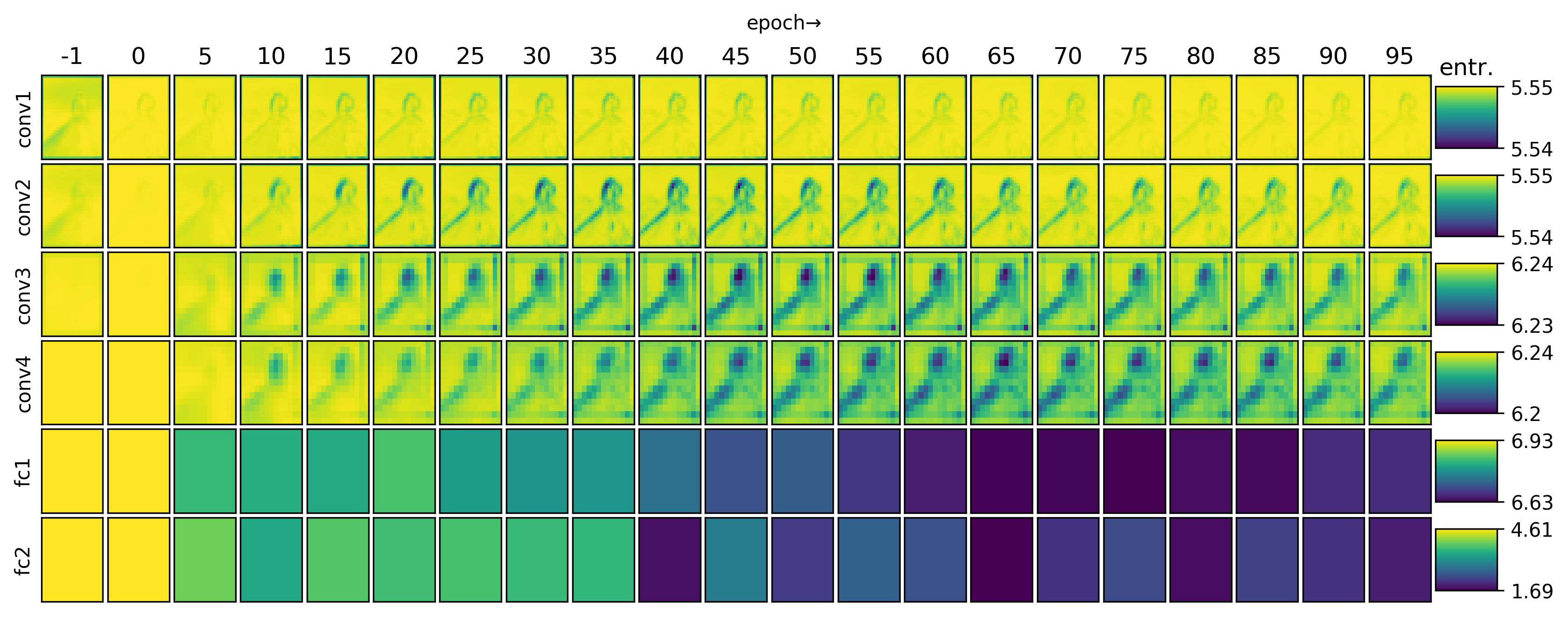}
            \end{minipage}
        }
    \end{minipage}

\caption{Sparsity visualization results. 
Resulting sparsity heatmaps for (a) vanillaNet, (b) dropNet, (c) normNet, (d) and sr1Net trainings.
The respective heatmaps were created before the training (epoch -1) and after each corresponding epoch.}
\label{fig:entropy-heatmaps}
\end{figure*}
\begin{figure*}
	\centering 
	\includegraphics[width=\linewidth]{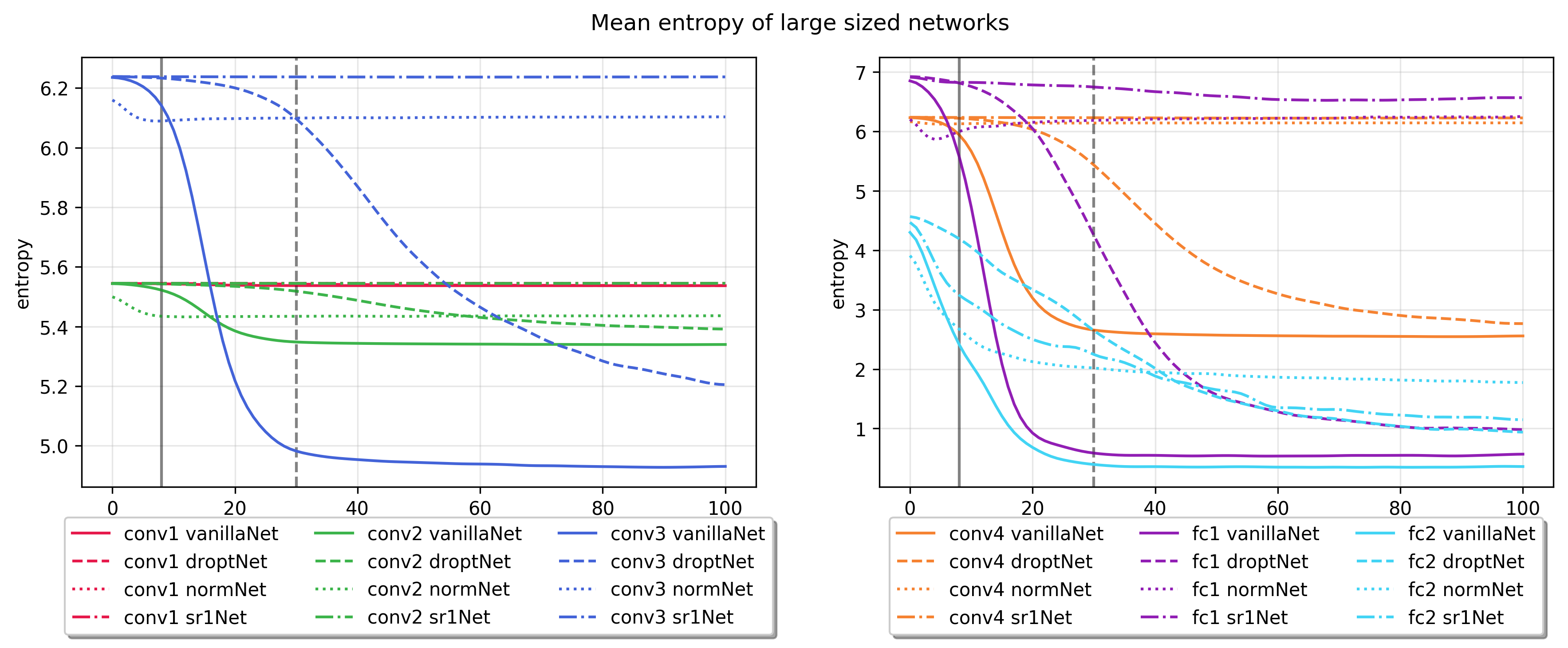}
	\centering
	\caption{Mean entropy of all RFAV of a given layer for unregularized, dropout, batch normalization and sparsity regularized (sr) trainings. The moment of overfitting (increase of validation loss) is highlighted by the vertical lines with corresponding line styles.
	Note that normNet and sr1Net do not suffer from overfitting so that no vertical lines are given for these architectures.}
	\label{fig:mean-entropy-xxl-unreg-drpt-bn}
\end{figure*}

\noindent Initially, small, medium-sized, and xxl vanillaNets were trained (see \autoref{table:network}). \autoref{fig:loss-all} depicts the evolution of their validation losses. 
All vanillaNet trainings show a significant increase of the validation loss values starting in early stages of training.
To visualize the localized sparsity of activations we are encoding the entropy of each proximate receptive field of a given layer in a heat map (see Figure~\ref{fig:entropy-heatmap:unreg}). 
Here, prominent features of the input image are recognizable by a lower entropy. 
Therefore, some filters of the respective layers generate a stronger activation for these features in comparison to the other filters. 
The visualized network starts overfitting from the 8th epoch (see Figure~\ref{fig:loss-large}, vanillaNet-xxl). 
In the heat maps, starting from the 10th epoch, a decrease of entropy  becomes apparent throughout the entire image in all layers. 
This change is analyzed in more detail in \autoref{fig:mean-entropy-xxl-unreg-drpt-bn}. 
Here, the mean of 1,000 entropy heat maps is plotted for all epochs. 
Especially the entropies of conv3, conv4, and fc1 undergo  a rapid change shortly before and after the moment of overfitting. 
As part of our experiments, we have been able to recognize this effect throughout the trainings.

\begin{table*}
\newlength{\colwidth}
\setlength{\colwidth}{1.1cm}
\centering
\begin{tabular}{ll|cccc|c|c}
\textbf{Name} & \textbf{Description}                   & \multicolumn{4}{c|}{\textbf{cifar-100}}                                   & \textbf{cifar-10} & \textbf{tiny}   \\ \hline
\textbf{}     & \textbf{}                              & \textbf{s}      & \textbf{m}      & \textbf{xxl}    & \textbf{xxl (Adam)} & \textbf{xxl}      & \textbf{xxl}    \\ \hline
vanillaNet    & No regularization                      & 0.3486          & 0.3603          & 0.3656          & 0.3843              & 0.6964            & 0.2198          \\
dropNet       & Dropout                                & 0.2505          & 0.4708          & 0.4202          & 0.4507                   & 0.7862                 & 0.2446          \\
normNet       & Batch normalization                    & 0.3626          & 0.4302          & 0.5059          & \textbf{0.5292}                   & 0.7900            & 0.2661          \\ \hhline{========}
sr1Net        & vanillaNet + SparsityReg (conv1 - fc1) & 0.2663          & 0.4917          & 0.5325          & 0.3960              & \textbf{0.8079}   & 0.2714          \\
sr2Net        & vanillaNet + SparsityReg (conv3 - fc1)  & 0.3460          & 0.5023          & 0.5387          & 0.4025     & 0.8042            & \textbf{0.3123} \\
sr3Net        & sr2Net + DecorrReg                     & \textbf{0.3781} & \textbf{0.5219} & \textbf{0.5423} & 0.4079              & 0.8040            & 0.3052         
\end{tabular}
\caption{Best network accuracy of different regularizations during 100 epochs of small, medium-sized and xxl network trainings on cifar-10, cifar-100 and tiny-imagenet (tiny).}
\label{table:accuracy-results}
\end{table*}
When applying dropout, overfitting in dropNet-s and dropNet-m is prevented over the entire course of training (see Figure~\ref{fig:loss-small} and~\ref{fig:loss-medium}). The dropNet-m and dropNet-xxl networks achieve a considerably lower loss than the corresponding vanillaNets. %
However, dropNet-xxl still overfits starting around epoch 30 (see Figure~\ref{fig:loss-large}). Again, the corresponding heat maps reveal a noticeable change in entropy around this moment of training (see Figure~\ref{fig:entropy-heatmap:drpt}). Similar to vanillaNet, distinct features are visible due to lower entropy but not as apparent as in its unregularized counterpart. The mean entropy plot also reveals a drop of entropy around the moment of overfitting, but not as strong as in vanillaNet (see \autoref{fig:mean-entropy-xxl-unreg-drpt-bn}). Nevertheless, dropNet shows a significantly higher performance for the xxl network training compared to vanillaNet (see \autoref{table:accuracy-results}). 

With the help of batch normalization, overfitting in the normNet-xxl network can be reduced. Nevertheless, the loss starts increasing slowly around the 20th epoch (see Figure~\ref{fig:loss-large}). In contrast to vanillaNet and dropNet, the sparsity heat maps of all layers in normNet remain stable except for the first epoch. The range of observed entropy values per layer remains small. %
The mean entropy hardly changes throughout the whole training (see \autoref{fig:mean-entropy-xxl-unreg-drpt-bn}). Compared to vanillaNet and dropNet, normNet achieves the best classification accuracy for all network sizes (see \autoref{table:accuracy-results}).

We observed that the use of dropout and batch normalization have almost always led to a lower RFAV sparsity and higher accuracies than the unregularized trainings (\textbf{C1}). Layer fc2 is not considered for the entropy analysis. Here, the entropy decreases in all trainings due the the use of the categorical cross-entropy loss function. A low entropy in the last fully-connected layer of a NN implies that the number of labels that are considered for classification is decreasing. 
This is an expected observation and is therefore not considered as related to the overfitting phenomenon here.   

\subsection{Targeted Sparsity Regularization}
\noindent In \autoref{sec:monitoring_sparsity}, 
we observed that overfitting can be associated with a strong change in RFAVs' mean entropy and therefore with a higher sparsity in the respective layers. 
To get a better understanding on how this affects our trainings, we apply SparsityReg to force the network to prevent sparsity and counteract overfitting (\textbf{C2}, \textbf{C4}).

First, the RFAV entropy of all layers is regularized (srNet1). The validation loss of these runs shows a decrease or convergence throughout the entire training (see \autoref{fig:loss-all}). The sr1Net-m and sr1Net-xxl trainings achieve a smaller loss than its  vanillaNet, dropNet, and normNet counterparts.
The accuracies of these regularized trainings can be seen in \autoref{table:accuracy-results}. 
We can see that the larger the network, the higher accuracies can be obtained. 
While this seems to be perfectly in line with expectations, it should be noted that vanillaNet and dropNet can not benefit from the increased size.
The sr1Net-xxl network achieves a significantly higher accuracy than the unregularized (+0.1669), dropout (+0.1123), and batch normalized (+0.0266) equivalents. 

Layer-wise analysis of entropy has revealed that especially the layers conv3, conv4, and fc1 show a strong change in entropy (see Figure~\ref{fig:mean-entropy-xxl-unreg-drpt-bn} and \textbf{C1}). When applying a targeted regularization to only these layers (sr2Net), validation loss reaches the lowest values across all network sizes. This targeted regularization also ensures that the networks do not overfit. Here, the accuracy is improved in nearly all trainings compared to sr1Net (see \autoref{table:accuracy-results}).

\begin{figure*}[t]
   \subfloat[\label{fig:correlation-histogram:unreg}]{
    \begin{minipage}[b]{0.44\textwidth}
       \includegraphics[width=\linewidth]{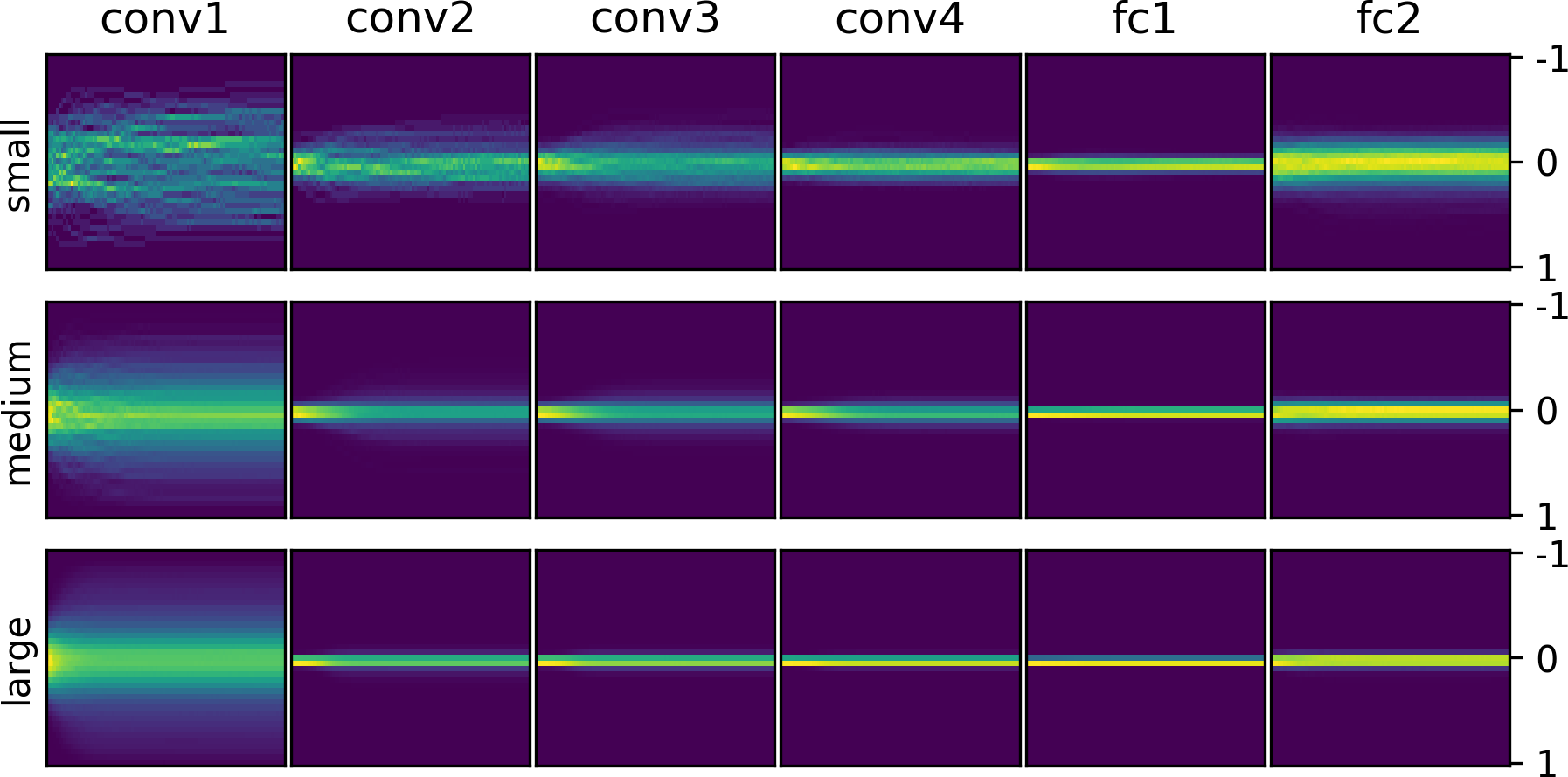}
       \end{minipage}
    }
    \hfill
    \hfill
    \begin{minipage}[b]{0.44\textwidth}
        \subfloat[\label{fig:correlation-histogram:sparse-reg}]{o
            \begin{minipage}[b]{\textwidth}
                \includegraphics[width=\linewidth]{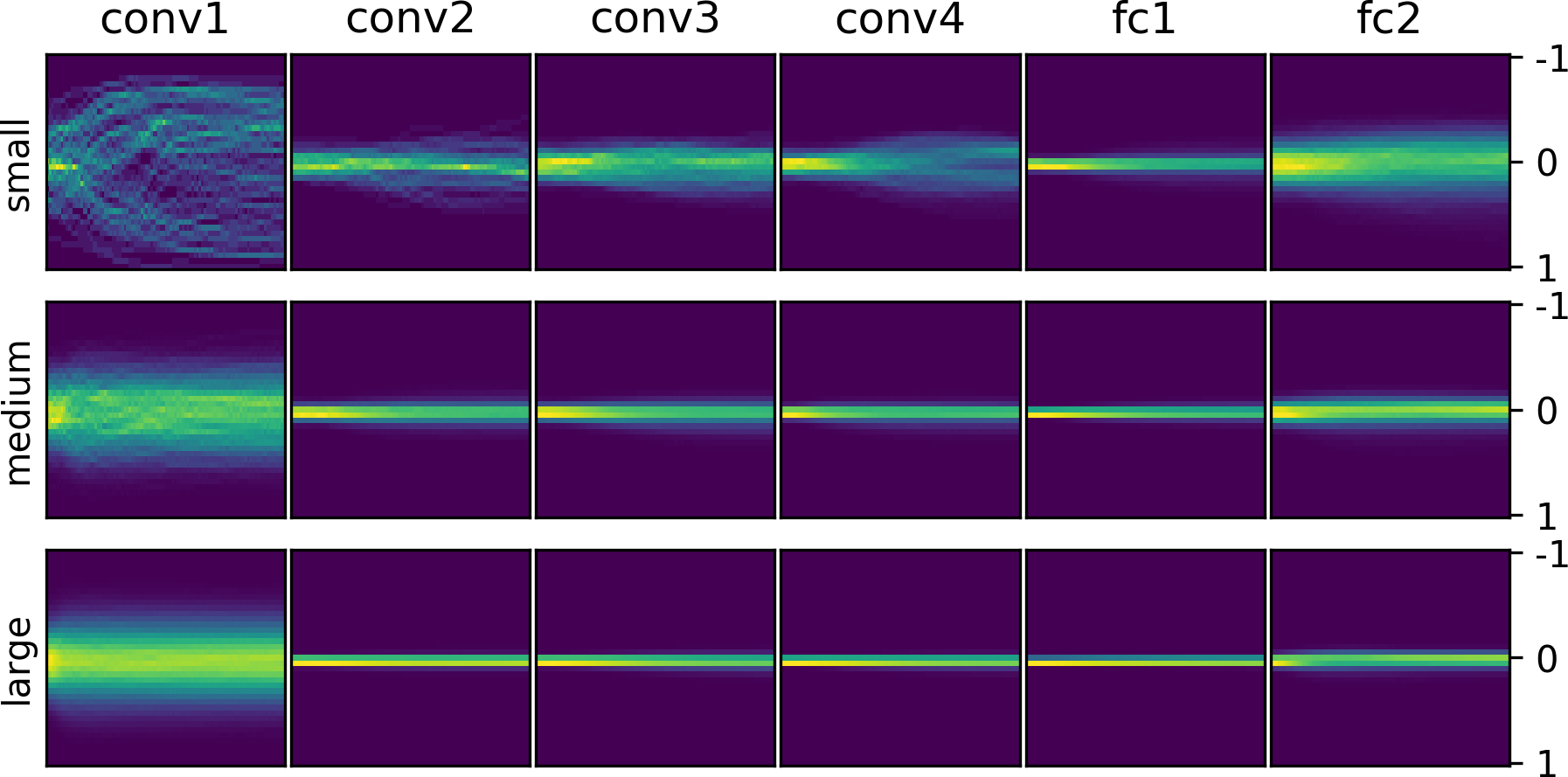}
            \end{minipage}
        }
    \end{minipage}
    \hfill
    \hfill
    \begin{minipage}[b]{0.098\textwidth}
        \subfloat[\label{fig:correlation-histogram:corr-reg}]{
            \begin{minipage}[b]{\textwidth}
                \includegraphics[width=\linewidth]{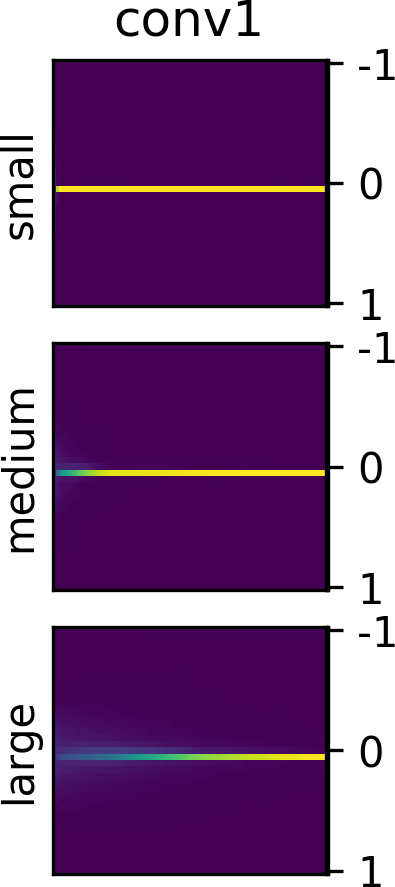}
            \end{minipage}
        }
    \end{minipage}
\caption{Histogram of neurons' correlation coefficients in individual layers over 100 epochs for vanillaNet (a) and sr1Net (b) trainings. Color encodes the frequency of how often correlation coefficient (rounded to two decimals) occur in the lower triangle of the correlation matrix from low (yellow) to high (blue). The $x$-axes of the respective histograms represent the corresponding epoch, the $y$-axes the correlation coefficients. The distribution of correlation coefficents of sr3Net's first layer is shown in (c), indicating a decorrelation of the conv1 layer compared to the corresponding layer given in (b).}
\label{fig:correlation-histogram}
\end{figure*}

In \autoref{sec:methods:regularizing-sparsity}, we pointed out that there is a risk of learning redundant filters when maximizing RFAV entropies. In order to improve the understanding of redundant filters, the pairwise correlation coefficients of the neuron weights of the respective layers are plotted in a histogram (see \autoref{fig:correlation-histogram} and \textbf{C1}). Comparing vanillaNet-xxl and sr1Net-xxl trainings, we observe that layers' correlations hardly differ. However, regardless of our regularization, the first layer stands out in all trainings by revealing a higher correlation among the individual neurons than in the other layers. Therefore, we used the DecorrReg on conv1 layer plus targeted SparsityReg (sr3Net) and achieved the highest accuracy throughout all experiments and network sizes on cifar-100 when trained for 100 epochs (see \autoref{table:accuracy-results}).
Furthermore, as shown in Figure~\ref{fig:correlation-histogram:corr-reg} the DecorrReg can effectively remove the correlations in the conv1 layer.

The analysis of sparsity-regularized losses has shown that even training large networks no longer exhibits an increase of validation loss (see Figure~\ref{fig:loss-large}). In another experiment, we tried to push sparsity-regularized networks into overfitting. Here, we trained the xxl networks for 200 epochs. It turned out that all runs, except for the sparsity-regularized trainings,  show a deterioration of the validation losses. All sparsity-regularized trainings, on the other hand, decrease continuously and converge against a certain value. Again, the sparsity-regularized runs achieved the lowest losses and the highest accuracies (up to 0.5513 by sr3Net).

\begin{figure}[t]
   \includegraphics[width=\linewidth]{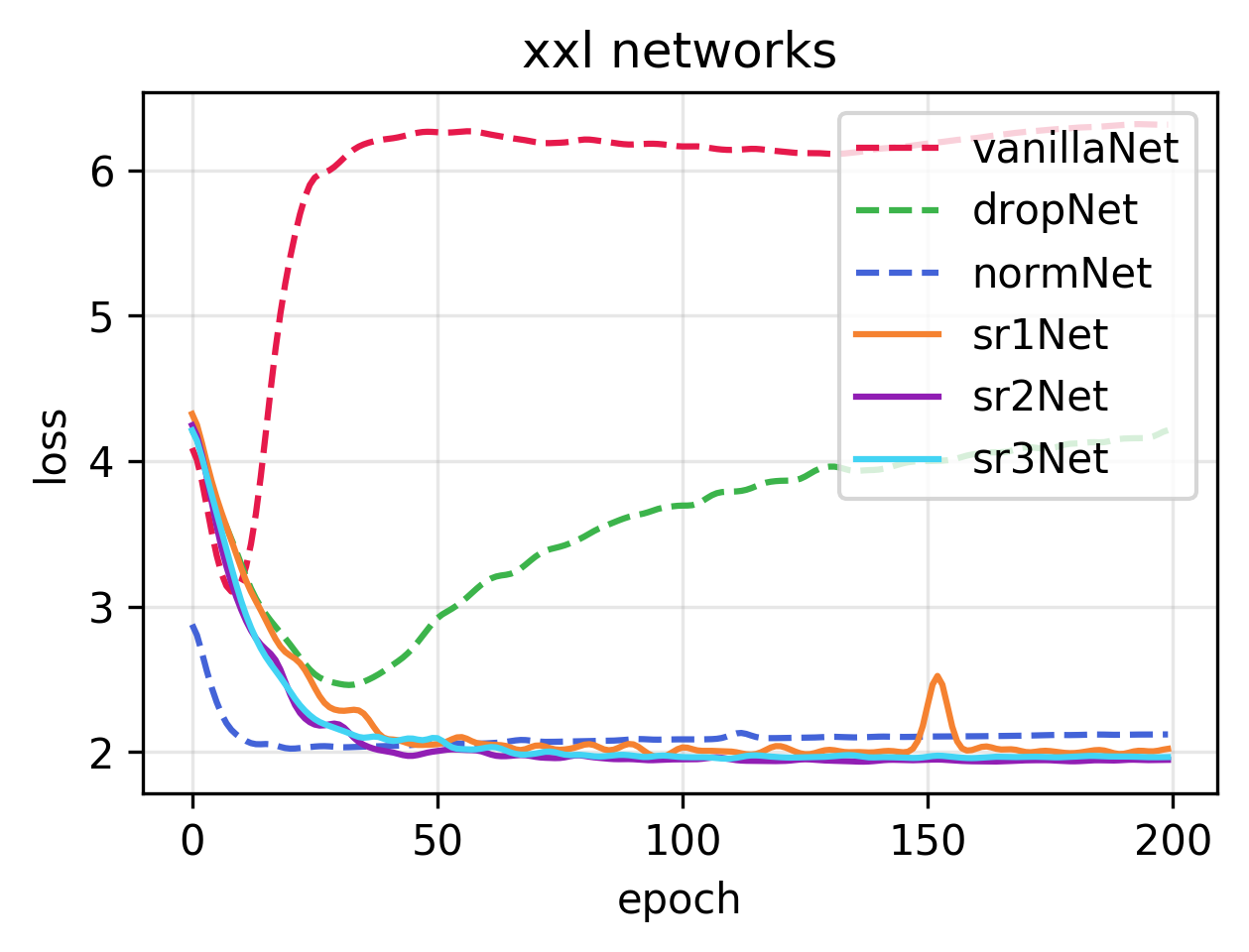}
\caption{Validation loss over 200 epochs of training of xxl networks. Except for sparsity-regularized trainings, all runs show an increasing loss throughout the course of training. }
\label{fig:loss-200-epochs}
\end{figure}

In addition to the cifar-100 dataset, we also trained the xxl networks on cifar-10 and tiny-imagenet. The classification accuracies can be seen in \autoref{table:accuracy-results}. 
Again, applying sparsity regularization we are able to outperform vanillaNet (+0.1115 / +0.0925), dropNet (+0.0217 / +0.0677), and normNet (+0.0179 / +0.0462) on cifar-10 / tiny-imagenet. 

Beside our baseline architectures, we also trained LeNet and AlexNet with and without sparsity regularization. LeNet training does not show a significant improvement of the classification accuracy. Our results so far have shown that sparsity regularization does not show any positive effects on small sized network trainings. Due to the size of the LeNet network, this can also be observed here. In contrast, AlexNet showed a significant improvement of the accuracy throughout all datasets when applying sparsity regularization.

As an alternative to SGD we also evaluated our regularization strategy against Adam optimization (\autoref{table:accuracy-results}).
In alignment with the reported tendency of this adaptive method \cite{yaguchi2018adam}, our visualization clearly indicates high sparsity (\autoref{fig:heatmap_adam}).
We further demonstrate that sparsity regularization can cause strongly correlating features if applied to method-induced sparsity (\autoref{fig:corr_hist_adam}). 
Interestingly, the effect of explicit decorrelation can counteract filter correlation but does not induce significant performance improvements during Adam optimization and thus requiring further investigations into the nature of sparsity and network capacity.
However, as shown in \autoref{table:accuracy-results}, our targeted regularization strategy achieves comparable performance on cifar-100 using a much smaller architecture (Adam xxl vs. sr3Net-m) and outperforms Adam with batch normalization if equally sized networks are used (Adam xxl vs. sr3Net-xxl).

\begin{table}
\begin{tabular}{l|ll|ll}
\multicolumn{1}{c|}{\textbf{Dataset}} & \multicolumn{2}{c|}{\textbf{\begin{tabular}[c]{@{}c@{}}Without\\ Regularization\end{tabular}}} & \multicolumn{2}{c}{\textbf{\begin{tabular}[c]{@{}c@{}}With\\ Regularization\end{tabular}}} \\ \hline
                                      & \textbf{LeNet}                                         & \textbf{AlexNet}                                        & \textbf{LeNet}                                        & \textbf{AlexNet}                                     \\ \hline
cifar-10                              & 0.6006                                        & 0.6987                                         & \textbf{0.6101}                              & \textbf{0.7509}                             \\
cifar-100                             & 0.2977                                        & 0.3842                                         & \textbf{0.3027}                              & \textbf{0.4682}                             \\
tiny-imagenet                         & 0.1267                                        & 0.2978                                         & \textbf{0.1345}                              & \textbf{0.3739}                            
\end{tabular}
\caption{Unregularized and sparsity-regularized LeNet and AlexNet trainings of cifar-10, cifar-100, and tiny-imagenet. }
\label{table:accuracy-architectures}
\end{table}

%% file: section/conclusion.tex
\section{Conclusion}
\noindent Throughout this paper we have studied sparsity-induced overfitting using novel visualizations. 
We conclude that sparse layer responses can be encoded by the  entropy of the receptive field activation vector. 
When overfitted, filters have been trained in a way that only few neurons have learned particular features of their respective inputs. 
These overconfident output distributions can be directly measured using our method. 
With the help of visualizations of the proximate receptive field entropy we are able to identify the layers in which the overfitting takes place and when it happens (\textbf{C1}). 
Sparsity heat maps are able to encode sparsity locally in the layers’ input space. 
Plotting the mean of several heat maps over all training epochs, we were able to identify which layers create overconfident responses when a network is overfitted. 

The analysis of proximate receptive field entropy has shown that the use of common regularizers such as dropout or batch normalization exhibit higher entropies compared to the unregularized counterpart for all layers. 
Instead of maximizing entropy implicitly by nesting additional layers to the network, we have developed a loss-based regularizer that explicitly maximizes the proximate receptive field entropy. 
With our novel regularization we are able to utilize the potential of large networks to learn cooperative %
features, pushing NNs to a higher generalization (\textbf{C4}). 
Applying our regularizers we are able to outperform otherwise identical dropout and batch-normalized networks  (\textbf{C3}).
As a result of our visualization, we are able to identify problematic layers in particular and thus apply regularizations in a targeted manner. 
By regularizing NN only where it is needed, we maintain the highest accuracies throughout all our experiments. 
Using our regularizer, we are able to avoid overfitting for more than 200 epochs on datasets and network sizes, in which their unregularized counterparts start overfitting before the 10th epoch. 

One potential risk of sparsity regularization (\ie rewarding high activations across all filters) is that all filters converge towards the most salient feature and thus maximize the entropy.
Using the cross-correlation between filters we could however demonstrate that targeted sparsity regularization (\ie reduced sparsity values) do not induce correlations across filters when trained with SGD. 
Nevertheless, we demonstrated that such side effects could be addressed by decorrelation regularization if necessary.
Moreover, DecorrReg has successfully been used to decorrelate the first layer which has further improved our network accuracy (cf. sr3Net).
In summary we showed that especially a combination of targeted sparsity and decorrelation regularizers prevents overfitting and outperforms all experiments (\textbf{C2}).

In future work, further experiments will be carried out which, for example, find the best possible hyperparameters for regularizations. The impact of different activation functions, optimizers and other common NN hyperparameters also needs more testing. 
Finally, we plan to test our methods on other tasks besides image classification.

%% file: section/appendix.tex
\clearpage
\newgeometry{top=20mm, bottom=40mm} 
\onecolumn 
\newpage
\appendix
\section{Supplementary Figures}
\setcounter{figure}{0} \renewcommand{\thefigure}{A.\arabic{figure}}
\begin{figure}[h]
\centering
\includegraphics[width=\linewidth]{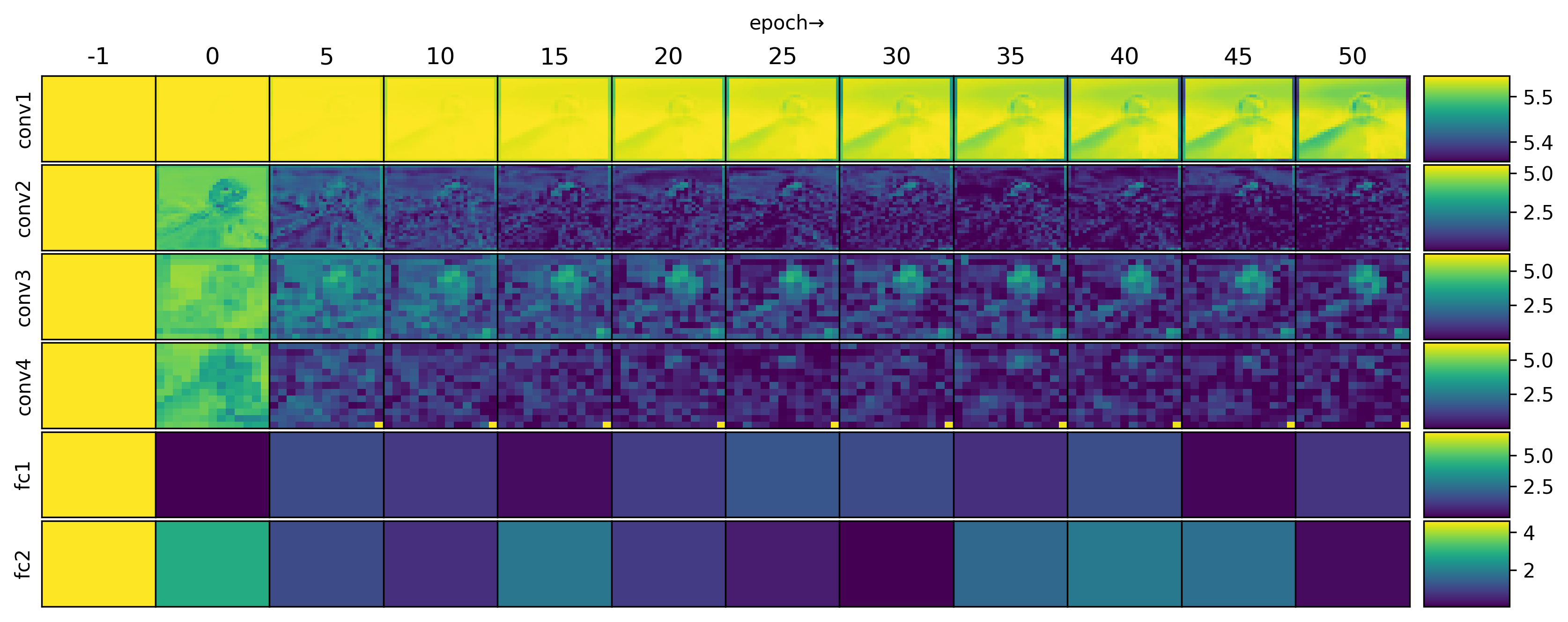}
\caption{Sparsity heatmaps of normNet using Adam optimizer (no targeted sparsity regularization was used). 
As visible from epoch 0 to 5 the entropy drops rapidly in the early stages of training resulting in high sparsity values.
These results confirm the findings reported in~[34] in which the authors demonstrate that Adam optimization induces sparsity in rectified networks.}
\label{fig:heatmap_adam}
\end{figure}
\begin{figure}[h]
\centering
\includegraphics[width=\linewidth]{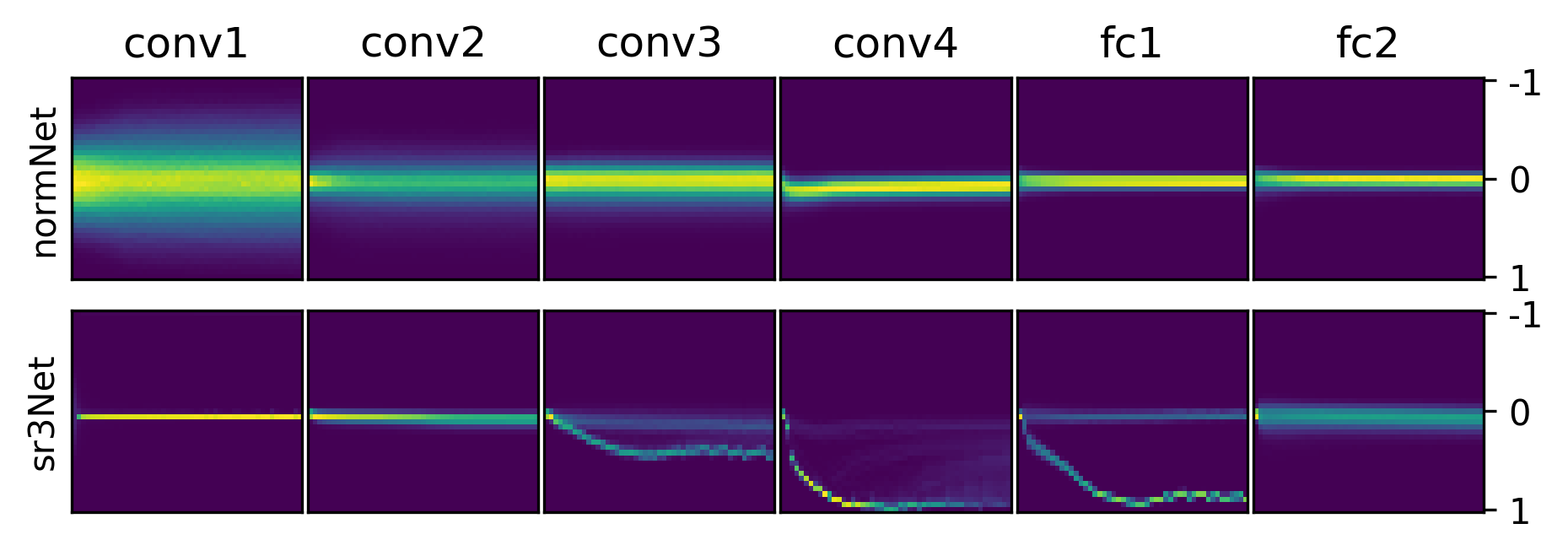}
\caption{Correlation histograms of normNet (without SparsityReg) and sr3Net (SparsityReg on conv3, conv4 and fc1 and DecorrReg on conv1) trainings using Adam optimizer. 
As can be seen the DecorrReg of conv1 efficiently removes the correlations visible in the corresponding normNet layer. 
In addition the filters of the sparsity regularized layers indicate strong correlation among neuron weights in the course of training.
This confirms the hypothesis that targeted sparsity regularization can induce filter correlations by maximizing the RFAV entropy.
}
\label{fig:corr_hist_adam}
\end{figure}